\documentclass[letterpaper]{article}
\usepackage{iccc}

\usepackage{times}
\usepackage{helvet}
\usepackage{courier}
\usepackage{graphicx}
\graphicspath{ {images/} }
\usepackage{subfig}
\usepackage{color}
\usepackage{url}
\title{Slogatron: Advanced Wealthiness Generator}
\author{Bryant Chandler and Aadesh Neupane}
\begin{document} 
\maketitle

\begin{abstract}

Creating catchy slogans is a demanding and clearly creative job for ad agencies. The process of slogan creation by humans involves finding key concepts of the company and its products, and developing a memorable short phrase to describe the key concept. We attempt to follow the same sequence, but with an evolutionary algorithm. A user inputs a paragraph describing describing the company or product to be promoted. The system randomly samples initial slogans from a corpus of existing slogans. The initial slogans are then iteratively mutated and improved using an evolutionary algorithm. Mutation randomly replaces words in an individual with words from the input paragraphs. Internal evaluation measures a combination of grammatical correctness, and semantic similarity to the input paragraphs. Subjective analysis of output slogans leads to the conclusion that the algorithm certainly outputs valuable slogans. External evaluation found that the slogans were somewhat successful in conveying a message, because humans were generally able to select the correct promoted item given a slogan.

\end{abstract}

\section{Introduction}
Slogans (also called taglines) are considered to be an important part of a company's identity and recognition. It provides the brand with its own unique image, which in turn helps in the brand recognition and recall in consumer's minds. Although they are vital for a company's brand value, there is very little consensus on what makes a good slogan. Researchers have identified various factors which contribute to make a good slogan. \cite{slogan_guidelines} reported that for a slogan to be successful it should be part of a strategic view of brand identity as it is capable of telling where the brand is going, and it must emphasize points of differentiation that are not only meaningful, but congruent with existing brand perception. Even given these attributes, some slogans fail to win the hearts of consumers. \cite{slogan_liking} found that the popularity for a slogan may be unrelated to media expenditure, and driven largely by the clarity of the message, the exposition of the benefits, rhymes, and creativity. Moreover, they found that jingles or brevity have no direct correlation on the like-ability of slogans.

Companies spend enormous sums of money to develop clever and impressive slogans, because slogans contribute to impression of a brand among people. The creation of a good slogan requires significant time, money and creativity for people experienced in that particular area. The creative team starts by researching the company's mission, vision, services, and projects. Then they start brainstorming about major topics related to the company. Lastly, they filter out slogans that don't fit the company's profile, and using trial and error, they select among the remaining lists of slogans.    

"Connecting people"- Nokia, "Diamonds are forever"- De Beers, "Melts in your mouth, not it your hands"- M\&M's are some examples of popular slogans. They are simple, consistent, highlight the company key characteristics, and provide credibility. Attempts have been made to embed these characteristics into creative computation models~\cite{brainsup,genetic_slogans}, but the results leave room for improvement.

Our system reads the company's mission, vision, and product description as input, and generates slogans using an evolutionary algorithm with a fitness function that attempts to favor the best slogans. The system outputs some of its highest internally rated slogans to the user. The goal is for the human to find slogans that they can use with little to no modification.

\section{Related Work}
Generation of creative artifacts has remained an interesting problem for computer scientists. They have experimented with machines which can create music, lyrics, visual art, paintings, recipes, jokes, poems, stories, news articles, games, and much more. We are most interested in studying machines that generate creative and catchy sentences.

\cite{segment_grammar} produced work about incremental sentence generation and described various constraints on the representation of the grammar for sentence generation. This generation method doesn't fall in the realm of creativity because of the strict use of a grammar, but it is an initial step toward sentence generation. The Poevolve~\cite{computational_poetic} system was able to generate poems using evolutionary techniques with neural networks for a fitness function, but the output was not syntactically cohesive. Poem generation can be complex, because it needs to follow many strict constraints such as: phonetics, linguistics, and semantics. \cite{computational_humour} developed a robust system for riddle/joke generation specifically designed for children with complex communication needs using the ideas from the JAPE~\cite{jape} system. It uses a large-scale multimedia lexicon and a set of symbolic rules to generate jokes.

Recently, many text generation methods have emerged to solve problems which are similar to our problem of slogan generation. One project~\cite{internet_memes} built a system capable of generating Internet Memes with coherent text, image, and humor values. Another~\cite{guerrero2015r} developed a twitter bot that generates riddles about celebrities by combining information from both well-structured and poorly-structured information sources. This~\cite{guerini2011slanting} system used WordNet~\cite{wordnet} and ConceptNet~\cite{conceptnet} to modify existing textual expressions to obtain more positively or negatively biased attitudes while keeping the original meaning. The modification is performed by adding or deleting words from existing sentences based on the their part-of-speech. Finally, \cite{slogans_news} blended well-known phrases with items from the news, in order to create interesting headlines. It was a worthy effort to create headlines in a linguistically motivated framework that accounts for syntactic aspects of language.

A precursor to the BrainSUP~\cite{brainsup} framework for slogan generation was a project working on creative naming~\cite{creative_naming}. It created a system which combines several linguistic resources and natural language processing techniques to generate neologisms based on homophonic puns and metaphors. It allowed users to determine the category of the service to be promoted together with the features to be emphasized. The researchers made use of WordNet and ConceptNet to get cognitively similar words.

Despite slogan generation being extremely socially relevant, the work already done in this sub-domain is limited. BrainSUP~\cite{brainsup} is a good attempt to accomplish creative sentence generation. It makes heavy use of syntactic information to enforce well-formed sentences, and constrain the search for a solution. Moreover, it allows users to force several words to appear in final generated sentences, and control the generation process across emotion, color, domain relatedness, and phonetic properties. The BrainSUP framework is general enough to be applied to other tasks. An extension of the framework is used in a successive study~\cite{keyword_evaluate}, which seeks to automate and evaluate the keyword method to teach secondary language vocabulary. Extrinsic evaluation confirmed the effectiveness of the proposed solution in the study.

This effort~\cite{genetic_slogans} is closely related to our application, as we were inspired by some of their approach to evolutionary slogan generation. Their method follows the BrainSUP framework for the initial population generation, and many of the evaluation schemes are also inspired from BrainSUP. The researchers have used evolutionary algorithms to generate slogans using eight evaluation functions to guide the generator towards best slogans. A new generation of slogans is created by pairing parents and performing crossover, followed by mutation with fixed probabilities. They have defined two varieties of crossover and mutation. Both varieties of crossover involve swapping different parts of a pair of parents in random locations based on POS tags. Mutation involves deletion of a random word, addition of an adverb, or replacement of a random word with its synonym, antonym, meronym, holonym, hypernym or hyponym.

{\bf Evolutionary Algorithms.}
Many solutions developed in the realm of computational creativity employ evolutionary algorithms for the generation of creative artifacts. One recipe generator~\cite{soup_recipe} used an evolutionary algorithm acting on a population of recipes, each composed of a list of ingredients to produce novel recipes. Another~\cite{manurung2012using} used evolutionary algorithms to generate meaningful poetry with sophisticated linguistic formalism for semantic representations and patterns of stress. The popularity of evolutionary algorithms in the field of computational creativity is due to its general architecture for generation and simple but effective mechanisms to control and steer the generation process.

There are too many ideas within the domain of evolutionary algorithms to enumerate, but we will review a few that have been used in this work. The first is mutation, which loosely encapsulates the idea of randomly modifying an individual within the population in some way. These mutations may or may not improve the performance of an individual. The second is niching or speciation, which allows an evolutionary algorithm to optimize for a problem with multiple solutions~\cite{mahfoud1995niching}. The approach allows multiple populations to develop independently, which provides populations that have optimized for different factors. As an addition to niching, occasional crossover can be done between niches in an attempt to produce even stronger individuals. The third is fitness function, which guides the algorithm towards producing refined individuals. The fitness function is a objective or constraint function which measures how well a given solution achieves the objective.

\section{Approach}
\begin{figure}[htb]
\centering
\includegraphics[width=\linewidth]{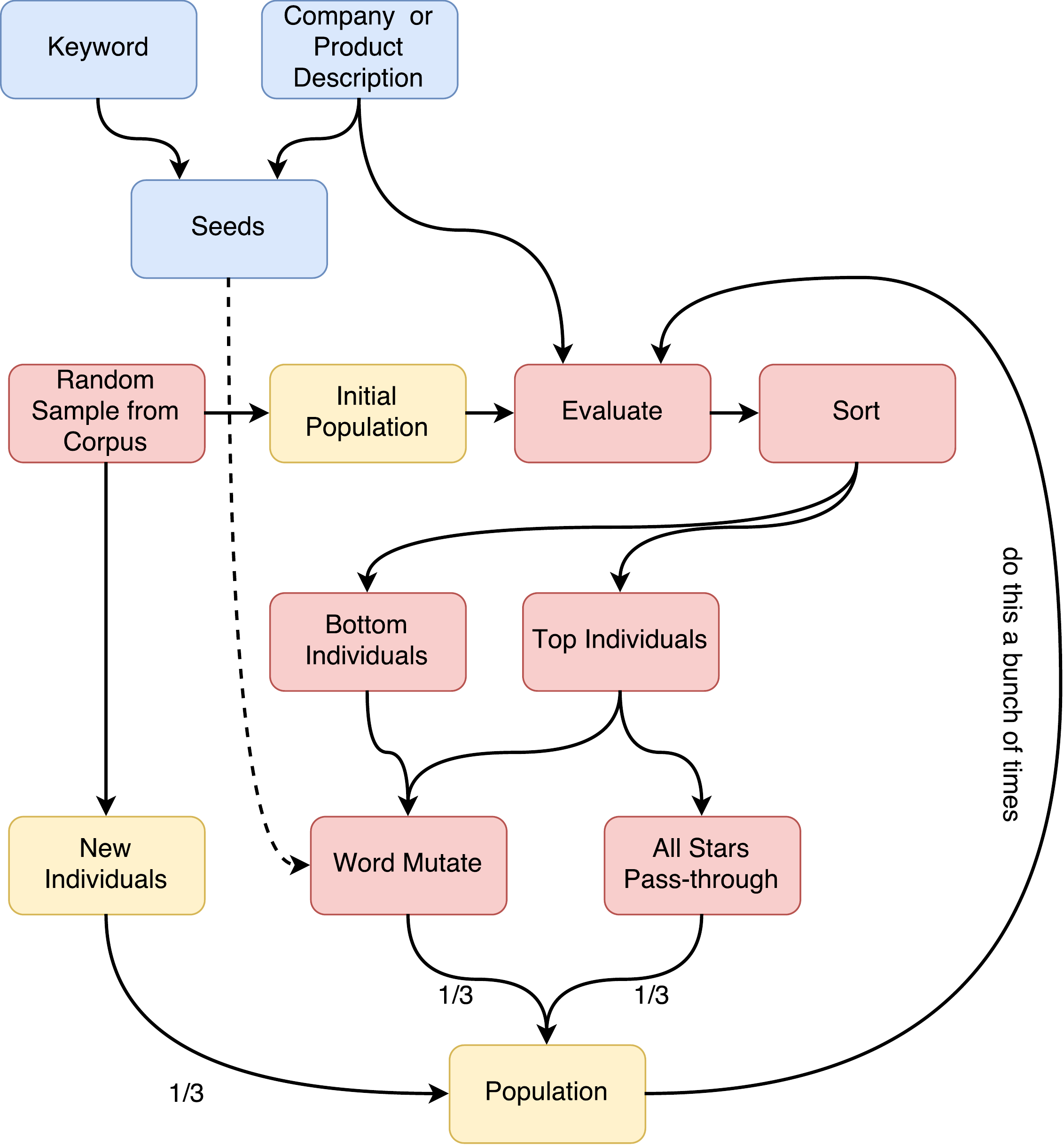}
\caption{System diagram}
\label{fig:system_diagram}
\end{figure}

Our solution to the slogan generation problem is based around a naive evolutionary algorithm with a layout as seen in Figure~\ref{fig:system_diagram}. The population for each generation after the initial generation is made up of $1/3$ newly sampled individuals, $1/3$ mutated individuals ($1/6$ from the top scoring and $1/6$ from the bottom scoring), and $1/3$ unmodified top scoring individuals. In this section we will discuss initial individual generation, mutation, and evaluation. Finally, we will discuss our use of speciation/niching to generate a variety of slogans.

\subsection{Initial Generation}   

For our system to generate final slogans, it needs an initial population. In an effort to give flexibility, we initially attempted to generate text using RNNs and Markov text generation. We found that it was difficult to generate satisfactory text using these methods. They can frequently repeat patterns, and the generated text is often not grammatically correct. Instead, we decided to initiate individuals by sampling from a corpus of existing text. We started by using Wikipedia and the Brown University corpus, but we found that the sampled sentences were generally overly complex and lacked a slogan-like structure. Given these issues, we moved to sampling from a corpus of almost 6000 slogans~\cite{wordlab}. Using this approach creates better output, but it has the downside of being highly correlated with existing slogans. Future work should look at ways to break out of the restriction to slogans that look like existing slogans.

\subsection{Mutation}
The goal of mutation is to inject words related to the seed summary into the slogan. During initialization of the algorithm we extract all nouns and verbs from the summary, and use the Datamuse API to get related nouns and verbs. From the related words we construct lists of seed nouns and verbs. Mutations are applied to the top $1/6$ and bottom $1/6$ of individuals. A mutation is performed by selecting a random word from the individual and replacing it with one of the seed words that belongs to the same part of speech. A weakness of this approach is that the conjugation of verbs might not match the original, which could make the individual have incorrect grammar. Future work should adjust conjugation before replacement.



\subsection{Evaluation}
Internal evaluation is important for the algorithm to filter out individuals that don't possess the desired characteristics and promote exemplary individuals. It is frequently represented as a fitness function in evolutionary algorithm literature. We split the evaluation function into englishness and semantic similarity. The score for each part is a number between zero and one, with the total score being a weighted sum of the two parts. Semantic similarity was weighted much less than englishness, because weighting it highly cause the algorithm to output slogans with too many of the seed words included. A low weight on semantic similarity allows the algorithm to favor applicable slogans, while still having good English.
 
{\bf Englishness.}
As described earlier, slogans have different properties that make them memorable and distinctive. Slogans with proper grammatical structure will be easier to remember and more attractive. For the algorithm to make this distinction, we have implemented an Englishness evaluation function using several techniques and variations. In the results we will discuss their strengths and weaknesses.

\begin{figure}[htb]
\centering
\includegraphics[width=\linewidth]{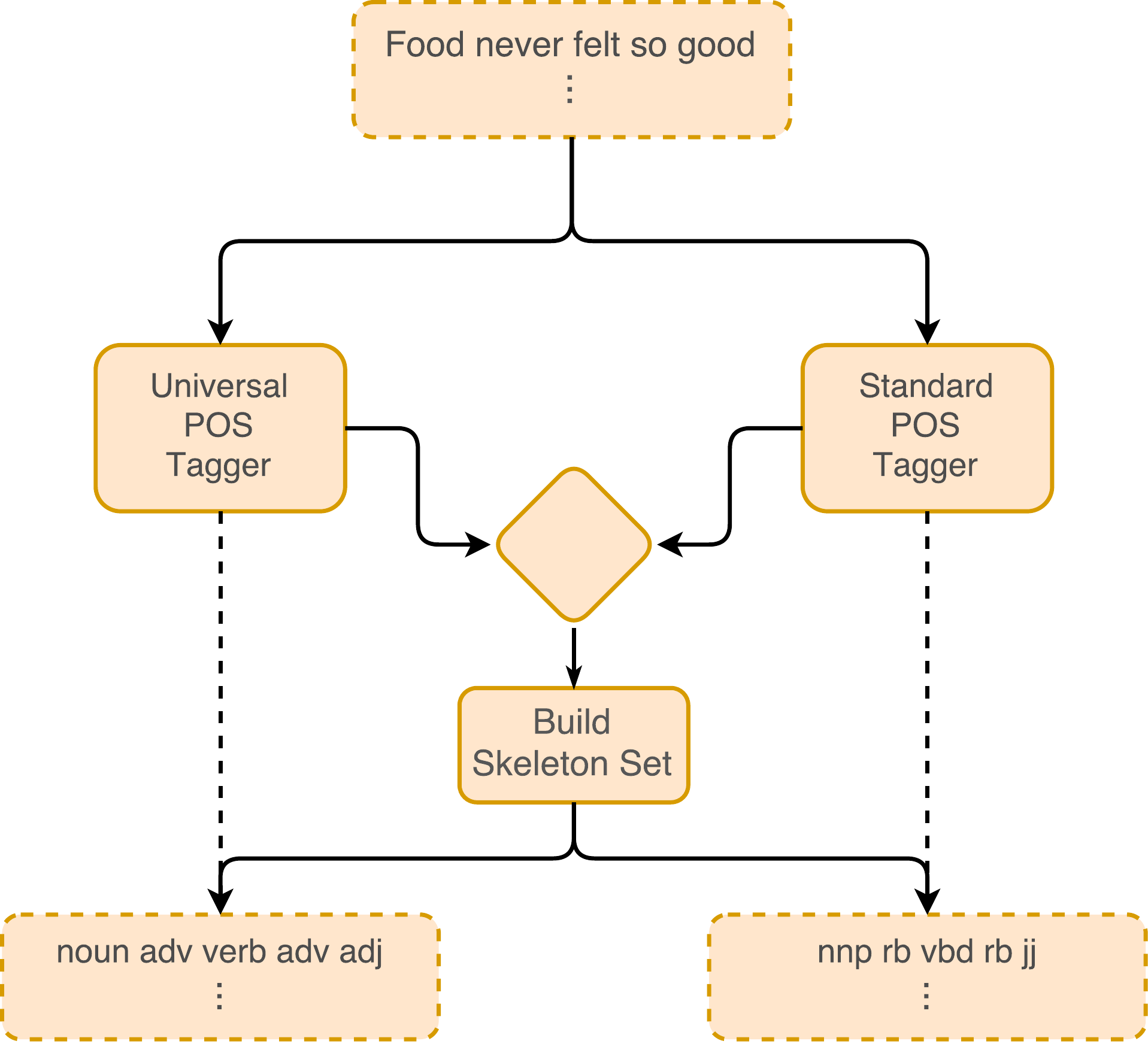}
\caption{Corpus is converted to POS Skeletons}
\label{fig:pos_skeleton}
\end{figure}

Several of the evaluation techniques use part of speech (POS) skeletons. A single POS skeleton is the list of part of speech tags for a single slogan. Skeletons are generated for each slogan in the corpus, and the set of all unique skeletons becomes the skeleton set as seen in Figure~\ref{fig:pos_skeleton}. Initially we used a standard POS tagger defined by the Penn Treebank Project, which has 36 tags. We hypothesized that using 36 tags would be too restrictive, so we also experimented with the universal POS tagger, which only has 12 tags. The idea is that using more general tagging will allow more interesting mutation.

The english measures were implemented as follows, with each measure being run with both standard and universal POS taggers:
\begin{itemize}
  
  \item {\bf POS Skeleton}: The scoring for this method is naive and very restrictive. An incoming individual is converted to a POS skeleton, and the algorithm checks if the skeleton is contained in the skeleton set. If individual is in the skeleton set, then it receives a score of one. If the individual isn't in the skeleton set, it receives a score of zero. The strength of this approach is that is frequently returns slogans with pleasing englishness. The main weakness is that the slogans obtained using this approach are frequently highly correlated with slogans from the inspiring set, with only one or two words replaced.

  
  \item {\bf N-Grams}: Initially, we downloaded the dataset of one million most frequent 2-grams and 3-grams in the Corpus of Contemporary American English (COCA). We made an assumption that the slogans containing those 2-grams and 3-grams would be more likely to be syntactically and semantically coherent. We counted how many 2-grams and 3-grams from the slogan appeared on the n-gram corpus, and took the ratio of count to length of sentence. The score is between zero and one, where values toward one denote that all of the n-grams appear in n-gram corpus and values towards zero signal that many n-grams are not in the corpus. Unfortunately, a corpus with only one million 2-grams is very small and we found that the uncommon words in slogans were unlikely to exist in the corpus, leading to incorrectly low scores. To combat this problem, we used a hybrid of the skeleton and n-gram approaches. We computed a set of 3-grams from the skeleton set, and computed scores based on 3-grams from the slogan that exist in the skeleton set as described previously. Later we will discuss the effect that this more flexible englishness measure has on the output.
\end{itemize}

{\bf Semantic Similarity.}
Humans are very effective at assessing the similarity of two sentences, but machines struggle to find very basic relations between sentences. Finding similarity between strings is a hard problem in NLP. WordNet, ConceptNet and skip-thought vectors~\cite{skip_vectors} are some projects, that work toward solving this problem. We are using cortical.io's Retina engine for semantic similarity evaluation function to measure the similarity of produced slogans and input text. The Retina engine~\cite{semantic_folding} is based on semantic folding, a novel representation of natural language based on sparse topological binary vectors. It claims to overcome the limitations of word embedding systems which use dense algebraic models, such as Word2Vec~\cite{word2vec} or Glove~\cite{glove} . Given two strings as input, the Retina engine outputs four different metrics for semantic similarity which are as follows:

\begin{itemize}
\item {\bf Jaccard Distance}: Measures dissimilarity between sample sets. Values close to zero mean that the strings are similar, whereas values close to one mean they are dissimilar.
\item {\bf Cosine Similarity}: Measures the similarity between samples purely based on word counts. Higher values indicate similarity.
\item {\bf Eculidean Distance}: Measures the similarity between samples. The lower the value the more similar, the samples are.
\item {\bf Weighted Scoring}: A weighted combination of the other three similarity scores. The API isn't clear about exactly how they are weighted.
\end{itemize}

This project used the weighted scoring, because it seemed to be the most generic similarity metric. Future work should explore other ways of measuring semantic similarity.

\subsection{Niching}
In evolutionary algorithms it is common for a population to converge to a single solution, with each individual only being a slightly different version of the same thing. For many applications this outcome would probably be acceptable, but we are interested in generating a variety of slogans to inspire the user. We use naive niching with no crossover. The different species never mingle, so the evolutionary algorithm can simply be run multiple times with the same input. The top individuals from each species are given as output. The approach is simple, but it greatly widens the range of output slogans.

\section{Results}
Most the analysis in this section will be subjective, and we encourage the reader to develop their own opinion about the value of the output. In addition to subjective analysis, we administered a survey to assess the external applicability of our outputs. 

\subsection{Englishness}
For the comparison of Englishness measures, we ran the algorithm once for each measure with the same input and selected a few top results. The item to be promoted was Hogwarts, and the input summary was, ``We educate young minds in the practice of witchcraft and wizardry for the service of others.''

{\bf Full skeleton with standard POS:}
\begin{itemize}
\item The magic of witchcraft
\item Righteous practice. Witchcraft
\item Magic is everyone’s witchcraft
\item Wizardry help and practice
\end{itemize}
The above results are nice slogans for Hogwarts. They seem to promote the school in a way that fits the input, and they have good grammar. A downside of these slogans is that they are all short. This might be a side-effect of the very restrictive englishness measure. It is much more likely for a short slogan to pass the test, so those are the only ones that make it through.

{\bf Full skeleton with universal POS:}
\begin{itemize}
\item Technical witchcraft for everyone
\item A witchcraft to get out of the magic
\item The sheer magic of practice
\item Exercise sorcery witchcraft
\end{itemize}
We again find respectable results for this approach. They promote Hogwarts while having correct grammar. The downside is again that they are all short.

{\bf Skeleton n-grams with standard POS:}
\begin{itemize}
\item Sorcery over minds
\item Magic of the practice
\item When the practice goes crazy, cultivate like wizardry. 
\item Genius witchcraft
\end{itemize}
Trade-offs begin to come into play with these n-gram results. The flexibility of the approach shows in the variety of outputs, but there is a reduction in englishness. Such a trade-off might be worthwhile if we use the system as inspiration for new slogans, rather than using the slogans without modification.

{\bf Skeleton n-grams with universal POS:}
\begin{itemize}
\item Natural practise and practice in your wizardry!
\item Come sorcery feeling like you’ve had a real witchcraft
\item One brain at a witchcraft
\item Acceptance educate the magic of the witchcraft
\end{itemize}
The englishness of the above sentences is certainly poor when compared with the other englishness measures, while the outputs certainly have more variety. The poorness of the English likely makes this approach unusable.

\subsection{External Evaluation}
For external evaluation we created a survey that provided the respondent with a generated slogan and asked them to select the item being promoted from four different options. Incorrect options in the were chosen in an effort to find middle ground between being too easy, and deliberately tricking the respondent. This approach certainly has some bias and should be viewed as anecdotal rather than scientifically conclusive. Future might include a free response survey so as to remove bias.

\begin{figure}[htb]
\centering
\includegraphics[width=\linewidth]{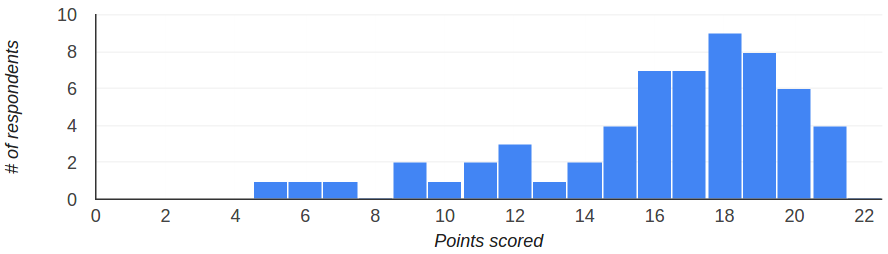}
\caption{Total points distribution}
\label{fig:point_distribution}
\end{figure}

The survey had 22 slogans generated from a variety of inputs, with a description at the top of the survey explaining the fact that the slogans were generated by an algorithm called Slogatron. There were 59 respondents with a median score of 17, a low of 5, and a high of 21. As can be observed in Figure~\ref{fig:point_distribution}, most respondents did very well, with only a few scoring very poorly.

As the results are anecdotal, we will discuss only a few of the slogans that had interesting results. The majority of slogan/item pairings were identified correctly by most respondents.

\begin{figure}[htb]
\centering
\includegraphics[width=\linewidth]{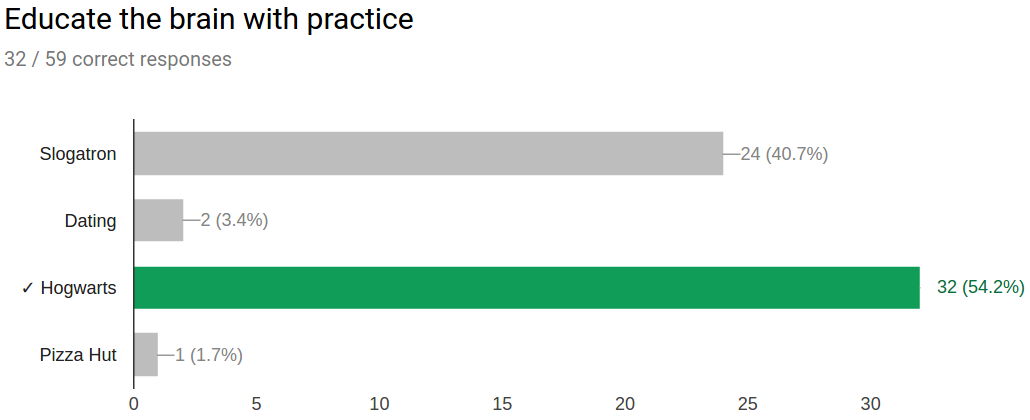}
\caption{Split between Hogwarts and Slogatron}
\label{fig:educate_with_practice}
\end{figure}

In Figure~\ref{fig:educate_with_practice} we see that respondents had trouble deciding if the slogan was about Slogatron or Hogwarts. We hypothesize that this is because people have some domain knowledge about Hogwarts, but only a cursory knowledge about Slogatron. The genericness of the slogan might have caused the respondents to incorrectly choose Slogatron.

\begin{figure}[htb]
\centering
\includegraphics[width=\linewidth]{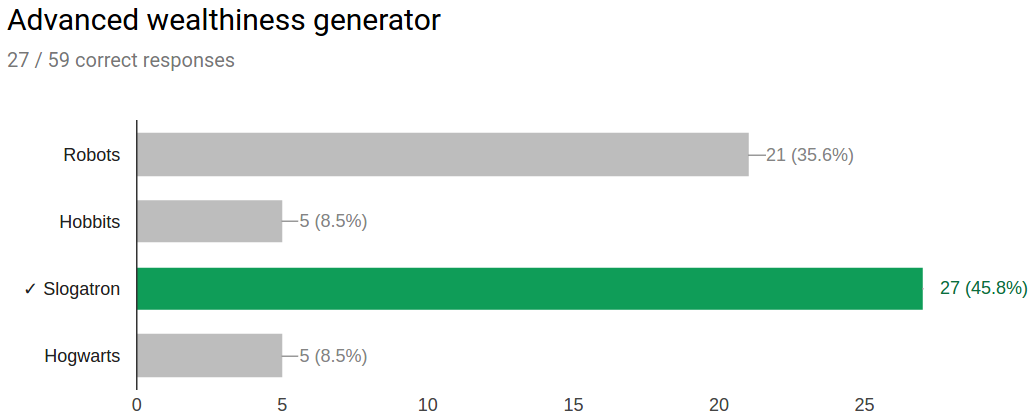}
\caption{Split between Robots and Slogatron}
\label{fig:advanced_wealthiness}
\end{figure}

Figure~\ref{fig:advanced_wealthiness} is another example of the respondents being obviously confused. Again, we hypothesize that this has something to do with the fact that people have a lot of prior information about robots, but not much about Slogatron.

\begin{figure}[htb]
\centering
\includegraphics[width=\linewidth]{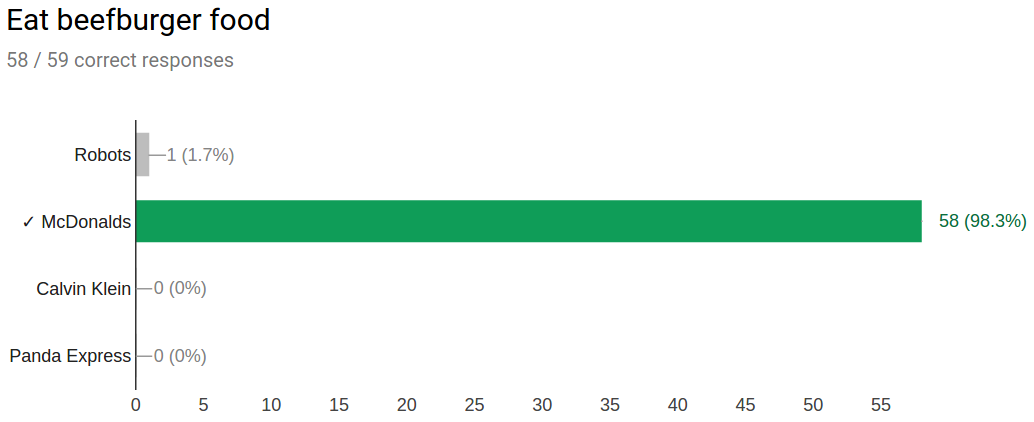}
\caption{Win for McDonald's}
\label{fig:beefburger}
\end{figure}

While none of the slogans had 100\% correct responses, Figure~\ref{fig:beefburger} shows a slogan for McDonald's with only one incorrect response. The success of this slogan is likely due to its semantic relatedess to the subject, and the widespread knowledge of McDonald's.

\begin{figure}[htb]
\centering
\includegraphics[width=\linewidth]{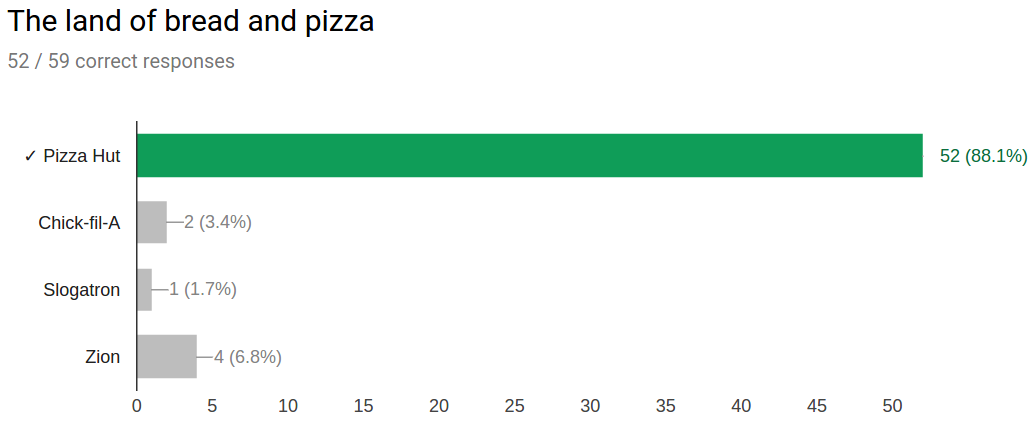}
\caption{Our favorite slogan}
\label{fig:bread_and_pizza}
\end{figure}

Finally, our favorite slogan from this experiment is found in Figure~\ref{fig:bread_and_pizza}. There was some minor dissent among respondents, but most agreed, correctly, that the slogan was promoting Pizza Hut. We like this output because it cleverly mutates the phrase, ``The land of milk and honey,'' to ``The land of bread and pizza.''

\section{Conclusions and Future Work}
 
While there is room for more scientific analysis of the slogans output from our algorithm, the outputs are subjectively pleasing. This algorithm could reasonably be used to inspire users toward creative slogans, and in many occasions it can provide slogans that are ready to use without modification.

We argue that our system is creative based on the creativity definition as provided by~\cite{framework_creative}, which states the a computer's output is creative if the same kind of output from a human would be considered creative. On the scale provided by~\cite{ventura2016mere}, we evaluate our system to be at the filtration level, as the system has self-evaluation mechanisms built-in.

The most critical improvements would be to englishness metrics, and perhaps in obtaining a larger corpus of slogans. While we were able to create slogans with good English, they seem to be restricted to short length, and are frequently highly correlated to the inspiring set. Eliminating these two weaknesses would go a long way toward increasing the creativity of the algorithm, and the output slogans would likely have more variety, which has often been called the spice of life.

An improvement to evaluation might come in the form of sentiment analysis. The idea is to find the sentiment of the input description given by the user and find the sentiments of each slogan from the system. A score of one is given if the sentiment of both input description and slogan match, else zero is given as the score. The enhanced Naive Bayes model~\cite{sentiment_analysis} might be a good model to use.

Future work should also include an expansion of the mutation step. Many good slogans include things such as metaphor and clever rhyming. A mutation step capable of injecting those components would likely dramatically increase the value of slogans produced.

In our experiments we found that the output slogans were heavily influenced by the input summaries. For example, longer summaries generally led to worse output. Future work should evaluate the benefits of applying a summarizer to pre-process the user input, possibly leading to more consistent results.

We found that many slogans were very close to making sense, but they had small conjugation or other grammatical flaws. We attempted to implement some auto-correction as found in~\cite{genetic_slogans}, but we weren't able to find a correction tool that functioned satisfactorily. Future work should include writing a correction tool from scratch that works with the specific needs of slogan generation.

\bibliographystyle{iccc}
\bibliography{iccc}

\begin{thebibliography}{}

\bibitem[\protect\citeauthoryear{Costa, Oliveira, and
  Pinto}{2015}]{internet_memes}
Costa, D.; Oliveira, H.~G.; and Pinto, A.~M.
\newblock 2015.
\newblock “in reality there are as many religions as there are
  papers”--first steps towards the generation of internet memes.
\newblock In {\em Proceedings of the Sixth International Conference on
  Computational Creativity June},  300.

\bibitem[\protect\citeauthoryear{Dass \bgroup et al.\egroup
  }{2014}]{slogan_liking}
Dass, M.; Kohli, C.; Kumar, P.; and Thomas, S.
\newblock 2014.
\newblock A study of the antecedents of slogan liking.
\newblock {\em Journal of Business Research} 67(12):2504--2511.

\bibitem[\protect\citeauthoryear{De~Smedt and Kempen}{1991}]{segment_grammar}
De~Smedt, K., and Kempen, G.
\newblock 1991.
\newblock Segment grammar: A formalism for incremental sentence generation.
\newblock In {\em Natural language generation in artificial intelligence and
  computational linguistics}. Springer.
\newblock  329--349.

\bibitem[\protect\citeauthoryear{De~Sousa~Webber}{2015}]{semantic_folding}
De~Sousa~Webber, F.
\newblock 2015.
\newblock Semantic folding theory and its application in semantic
  fingerprinting.
\newblock {\em arXiv preprint arXiv:1511.08855}.

\bibitem[\protect\citeauthoryear{Gatti \bgroup et al.\egroup
  }{2015}]{slogans_news}
Gatti, L.; {\"O}zbal, G.; Guerini, M.; Stock, O.; and Strapparava, C.
\newblock 2015.
\newblock Slogans are not forever: Adapting linguistic expressions to the news.
\newblock In {\em IJCAI},  2452--2458.

\bibitem[\protect\citeauthoryear{Guerini, Strapparava, and
  Stock}{2011}]{guerini2011slanting}
Guerini, M.; Strapparava, C.; and Stock, O.
\newblock 2011.
\newblock Slanting existing text with valentino.
\newblock In {\em Proceedings of the 16th international conference on
  Intelligent user interfaces},  439--440.
\newblock ACM.

\bibitem[\protect\citeauthoryear{Guerrero \bgroup et al.\egroup
  }{2015}]{guerrero2015r}
Guerrero, I.; Verhoeven, B.; Barbieri, F.; Martins, P.; et~al.
\newblock 2015.
\newblock The $\{$R$\}$ iddler $\{$B$\}$ ot:$\{$A$\}$ next step on the ladder
  towards creative $\{$T$\}$ witter bots.

\bibitem[\protect\citeauthoryear{Kiros \bgroup et al.\egroup
  }{2015}]{skip_vectors}
Kiros, R.; Zhu, Y.; Salakhutdinov, R.~R.; Zemel, R.; Urtasun, R.; Torralba, A.;
  and Fidler, S.
\newblock 2015.
\newblock Skip-thought vectors.
\newblock In {\em Advances in neural information processing systems},
  3294--3302.

\bibitem[\protect\citeauthoryear{Kohli, Leuthesser, and
  Suri}{2007}]{slogan_guidelines}
Kohli, C.; Leuthesser, L.; and Suri, R.
\newblock 2007.
\newblock Got slogan? guidelines for creating effective slogans.
\newblock {\em Business Horizons} 50(5):415--422.

\bibitem[\protect\citeauthoryear{Levy}{2001}]{computational_poetic}
Levy, R.~P.
\newblock 2001.
\newblock A computational model of poetic creativity with neural network as
  measure of adaptive fitness.
\newblock In {\em Proceedings of the ICCBR-01 Workshop on Creative Systems}.

\bibitem[\protect\citeauthoryear{Liu and Singh}{2004}]{conceptnet}
Liu, H., and Singh, P.
\newblock 2004.
\newblock Conceptnet—a practical commonsense reasoning tool-kit.
\newblock {\em BT technology journal} 22(4):211--226.

\bibitem[\protect\citeauthoryear{Mahfoud}{1995}]{mahfoud1995niching}
Mahfoud, S.~W.
\newblock 1995.
\newblock Niching methods for genetic algorithms.
\newblock {\em Urbana} 51(95001):62--94.

\bibitem[\protect\citeauthoryear{Manurung, Ritchie, and
  Thompson}{2012}]{manurung2012using}
Manurung, R.; Ritchie, G.; and Thompson, H.
\newblock 2012.
\newblock Using genetic algorithms to create meaningful poetic text.
\newblock {\em Journal of Experimental \& Theoretical Artificial Intelligence}
  24(1):43--64.

\bibitem[\protect\citeauthoryear{Mikolov \bgroup et al.\egroup
  }{2013}]{word2vec}
Mikolov, T.; Chen, K.; Corrado, G.; and Dean, J.
\newblock 2013.
\newblock Efficient estimation of word representations in vector space.
\newblock {\em arXiv preprint arXiv:1301.3781}.

\bibitem[\protect\citeauthoryear{Miller}{1995}]{wordnet}
Miller, G.~A.
\newblock 1995.
\newblock Wordnet: a lexical database for english.
\newblock {\em Communications of the ACM} 38(11):39--41.

\bibitem[\protect\citeauthoryear{Morris \bgroup et al.\egroup
  }{2012}]{soup_recipe}
Morris, R.~G.; Burton, S.~H.; Bodily, P.~M.; and Ventura, D.
\newblock 2012.
\newblock Soup over bean of pure joy: Culinary ruminations of an artificial
  chef.
\newblock In {\em Proceedings of the 3rd international conference on
  computational creativity},  119--125.

\bibitem[\protect\citeauthoryear{Narayanan, Arora, and
  Bhatia}{2013}]{sentiment_analysis}
Narayanan, V.; Arora, I.; and Bhatia, A.
\newblock 2013.
\newblock Fast and accurate sentiment classification using an enhanced naive
  bayes model.
\newblock In {\em International Conference on Intelligent Data Engineering and
  Automated Learning},  194--201.
\newblock Springer.

\bibitem[\protect\citeauthoryear{{\"O}zbal and
  Strapparava}{2012}]{creative_naming}
{\"O}zbal, G., and Strapparava, C.
\newblock 2012.
\newblock A computational approach to the automation of creative naming.
\newblock In {\em Proceedings of the 50th Annual Meeting of the Association for
  Computational Linguistics: Long Papers-Volume 1},  703--711.
\newblock Association for Computational Linguistics.

\bibitem[\protect\citeauthoryear{{\"O}zbal, Pighin, and
  Strapparava}{2013}]{brainsup}
{\"O}zbal, G.; Pighin, D.; and Strapparava, C.
\newblock 2013.
\newblock Brainsup: Brainstorming support for creative sentence generation.
\newblock In {\em ACL (1)},  1446--1455.

\bibitem[\protect\citeauthoryear{{\"O}zbal, Pighin, and
  Strapparava}{2014}]{keyword_evaluate}
{\"O}zbal, G.; Pighin, D.; and Strapparava, C.
\newblock 2014.
\newblock Automation and evaluation of the keyword method for second language
  learning.
\newblock In {\em ACL (2)},  352--357.

\bibitem[\protect\citeauthoryear{Pennington, Socher, and Manning}{2014}]{glove}
Pennington, J.; Socher, R.; and Manning, C.~D.
\newblock 2014.
\newblock Glove: Global vectors for word representation.
\newblock In {\em EMNLP}, volume~14,  1532--1543.

\bibitem[\protect\citeauthoryear{Ritchie \bgroup et al.\egroup
  }{2007}]{computational_humour}
Ritchie, G.; Manurung, R.; Pain, H.; Waller, A.; Black, R.; and O’Mara, D.
\newblock 2007.
\newblock A practical application of computational humour.
\newblock In {\em Proceedings of the 4th International Joint Conference on
  Computational Creativity},  91--98.

\bibitem[\protect\citeauthoryear{Ritchie}{2003}]{jape}
Ritchie, G.
\newblock 2003.
\newblock The jape riddle generator: technical specification.
\newblock {\em Institute for Communicating and Collaborative Systems}.

\bibitem[\protect\citeauthoryear{Tomasic, Papa, and
  Znidarsic}{2015}]{genetic_slogans}
Tomasic, P.; Papa, G.; and Znidarsic, M.
\newblock 2015.
\newblock Using a genetic algorithm to produce slogans.
\newblock {\em Informatica} 39(2):125.

\bibitem[\protect\citeauthoryear{Ventura}{2016}]{ventura2016mere}
Ventura, D.
\newblock 2016.
\newblock Mere generation: Essential barometer or dated concept.
\newblock In {\em Proceedings of the Seventh International Conference on
  Computational Creativity, ICCC}.

\bibitem[\protect\citeauthoryear{Wiggins}{2006}]{framework_creative}
Wiggins, G.~A.
\newblock 2006.
\newblock A preliminary framework for description, analysis and comparison of
  creative systems.
\newblock {\em Knowledge-Based Systems} 19(7):449--458.

\bibitem[\protect\citeauthoryear{Wordlab}{2017}]{wordlab}
Wordlab.
\newblock 2017.
\newblock {Wordlab Taglines/Slogans}.
\newblock \url{http://www.wordlab.com/archives/taglines-slogans-list/}.
\newblock [Online; accessed 25-April-2017].

\end{thebibliography}

\end{document}